\documentclass[11pt,a4paper]{article}
\usepackage[hyperref]{naaclhlt2018}
\usepackage{times}
\usepackage{latexsym}
\usepackage{paralist}
\usepackage{enumitem}
\usepackage{verbatim}
\usepackage[normalem]{ulem}
\usepackage{dirtytalk}
\usepackage[toc,page]{appendix}
\usepackage{amsmath}
\usepackage{amssymb}
\usepackage{tabularx}
\usepackage{floatrow}
\usepackage{url}
\usepackage{array}
\usepackage{multirow}
\usepackage{rotating}
\usepackage{booktabs}
\usepackage{subcaption}
\usepackage{bibunits}
\defaultbibliography{naaclhlt2018}
\defaultbibliographystyle{acl_natbib}

\newcolumntype{L}[1]{>{\raggedright\let\newline\\\arraybackslash\hspace{0pt}}m{#1}}

\aclfinalcopy 

\def\aclpaperid{***} 

\title{Punny Captions: Witty Wordplay in Image Descriptions}

\author{Arjun Chandrasekaran$^{1}$ \qquad Devi Parikh$^{1,2}$ \qquad Mohit Bansal$^3$  \\
$^1$Georgia Institute of Technology \qquad $^2$Facebook AI Research \qquad  $^3$UNC Chapel Hill  \\
{\tt \small \{carjun, parikh\}@gatech.edu \qquad mbansal@cs.unc.edu}}
\date{}

\begin{document}
\maketitle
\begin{abstract}
  Wit is a form of rich interaction that is often grounded in a specific situation (e.g., a comment in response to an event). 
  In this work, we attempt to build computational models that can produce witty descriptions for a given image. 
  Inspired by a cognitive account of humor appreciation, we employ linguistic wordplay, specifically puns, in image descriptions. 
  We develop two approaches which involve retrieving witty descriptions for a given image from a large corpus of sentences, or generating them via an encoder-decoder neural network architecture.
  We compare our approach against meaningful baseline approaches via human studies and show substantial improvements.
  We find that when a human is subject to similar constraints as the model regarding word usage and style, people vote the image descriptions generated by our model to be slightly wittier than human-written witty descriptions. 
  Unsurprisingly, humans are almost always wittier than the model when they are free to choose the vocabulary, style, etc. 

\end{abstract}
\begin{bibunit}
\section{Introduction}
\begin{quote}
``\emph{Wit is the sudden marriage of ideas which before their union were not perceived to have any relation.}''
\end{quote}
\begin{flushright}
-- Mark Twain 
\end{flushright}

Witty remarks are often contextual, i.e., grounded in a specific situation. 
Developing computational models that can emulate rich forms of interaction like contextual humor, is a crucial step towards making human-AI interaction more natural and more engaging~\cite{yu2016wizard}. 
E.g., witty chatbots could help relieve stress and increase user engagement by being more personable and human-like. 
Bots could automatically post witty comments (or suggest witty responses) on social media, chat, or messaging. 

\begin{figure}[t!]
\setlength{\fboxsep}{0pt}
\setlength{\fboxrule}{0pt}
	\begin{subfigure}[t]{1.42in}
 \fbox{\includegraphics[width=1.4in,height=1.2in]{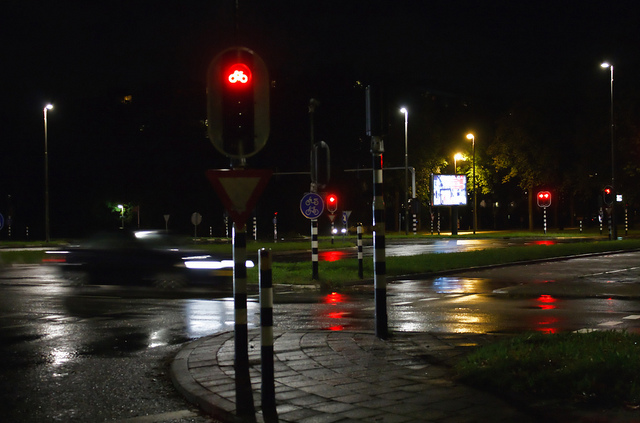}}
		\caption{\textbf{Generated}: a poll (pole) on a city street at night. \\ \textbf{Retrieved}: the light knight (night) chuckled. \\ \textbf{Human}: the knight (night) in shining armor drove away.}
        \label{fig:knight}
	\end{subfigure}
    \hfill    
	\begin{subfigure}[t]{1.55in}
  \fbox{\includegraphics[width=1\linewidth,height=1.2in]{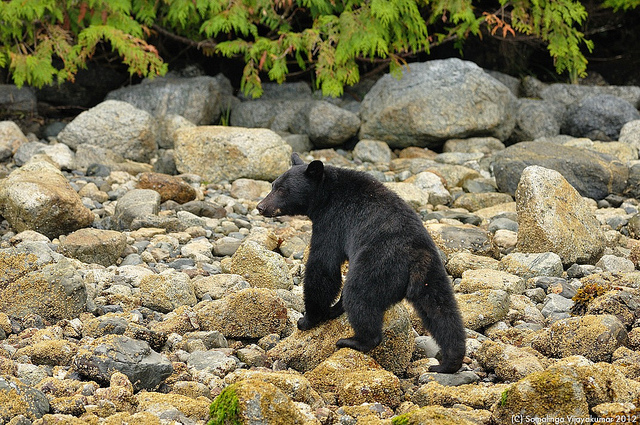}}
			\caption{\textbf{Generated}: a bare (bear) black bear walking through a forest. \\ \textbf{Retrieved}: another reporter is standing in a bare (bear) brown field. \\ \textbf{Human}: the bear killed the lion with its bare (bear) hands.}
        \label{fig:bear}
	\end{subfigure}
\caption{
\fontsize{10}{11}\selectfont{
Sample images and witty descriptions from 2 models, and a human. 
The words inside `()' (e.g., pole and bear) are the puns associated with the image, i.e., the source of the unexpected puns used in the caption (e.g., poll and bare).} 
}
\label{fig:teaser}
\end{figure}

The absence of large scale corpora of witty captions and the prohibitive cost of collecting such a dataset (being witty is harder than just describing an image) makes the problem of producing contextually witty image descriptions challenging. 

In this work, we attempt to tackle the challenging task of producing witty (pun-based) remarks for a given (possibly boring) image. 
Our approach is inspired by a two-stage cognitive account of humor appreciation~\cite{TwoStageModelHumor} which states that a perceiver experiences humor when a stimulus such as a joke, captioned cartoon, etc., causes an \emph{incongruity}, which is shortly followed by \emph{resolution}. 

We introduce an incongruity in the perceiver's mind while describing an image by using an unexpected word that is phonetically similar (pun) to a concept related to the image.
E.g., in Fig.~\ref{fig:bear}, the expectations of a perceiver regarding the image (bear, stones, etc.) is momentarily disconfirmed by the (phonetically similar) word `bare'. This incongruity is resolved when the perceiver parses the entire image description. 
The incongruity followed by resolution can be perceived to be witty.\footnote{Indeed, a perceiver may fail to appreciate wit if the process of `solving' (resolution) is trivial (the joke is obvious) or too complex (they do not `get' the joke).} 

We build two computational models based on this approach to produce witty descriptions for an image. 
First, a model that retrieves sentences containing a pun that are relevant to the image from a large corpus of stories~\cite{zhu2015aligning}. 
Second, a model that generates witty descriptions for an image using a modified inference procedure during image captioning which includes the specified pun word in the description. 

Our paper makes the following contributions: 
To the best of our knowledge, this is the first work that tackles the challenging problem of producing a witty natural language remark in an everyday (boring) context.
We present two novel models to produce witty (pun-based) captions for a novel (likely boring) image.  
Our models rely on linguistic wordplay. They use an unexpected pun in an image description during inference/retrieval. Thus, they do not require to be trained with witty captions. 
Humans vote the descriptions from the top-ranked \emph{generated} captions `wittier' than three baseline approaches.  
Moreover, in a Turing test-style evaluation, our model's best image description is found to be wittier than a witty human-written caption\footnote{
This data is available on the author's webpage.
} 55\% of the time when the human is subject to the same constraints as the machine regarding word usage and style.

\begin{figure*}[t!]
    \centering
    \includegraphics[scale=0.19]{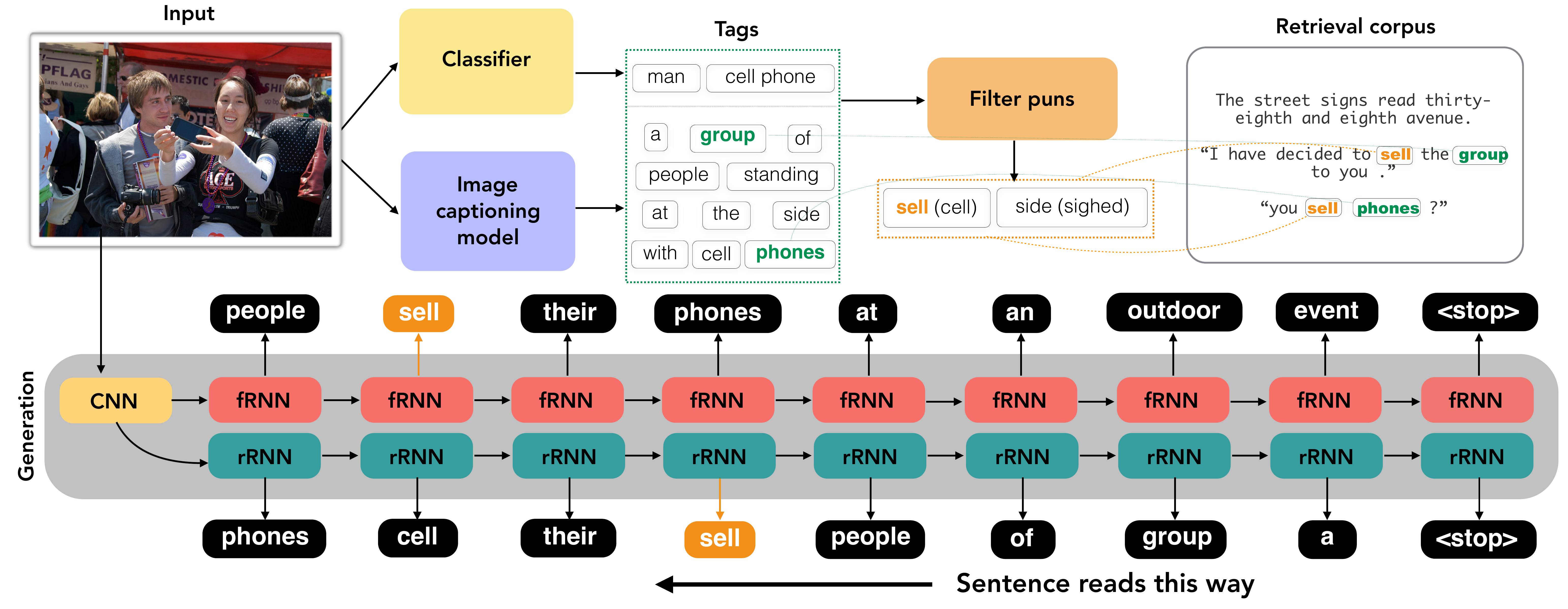}
    \caption{
       \fontsize{10}{11}\selectfont{Our models for generating and retrieving image descriptions containing a pun (see Sec.~\ref{sec:approach}).}} 
    \label{fig:model}
\end{figure*}
\section{Related Work}
\label{sec:relatedwork}

\noindent
\textbf{Humor theory.} General Theory of Verbal Humor~\cite{attardo1991script} characterizes linguistic stimuli that induce humor but implementing computational models of it requires severely restricting its assumptions~\cite{binsted1996machine}. 

\noindent 
\textbf{Puns.}~\newcite{zwicky1986imperfect} classify puns as \emph{perfect} (pronounced exactly the same) or \emph{imperfect} (pronounced differently). Similarly,~\newcite{pepicello1984language} categorize riddles based on the linguistic ambiguity that they exploit -- phonological, morphological or syntactic. 
~\newcite{Kao2013-kx} formalize the notion of incongruity in puns and use a probabilistic model to evaluate the funniness of a sentence.
~\newcite{Jaech2016PhonologicalP} learn phone-edit distances to predict the counterpart, given a pun by drawing from automatic speech recognition techniques. In contrast, we augment a web-scraped list of puns using an existing model of pronunciation similarity. 

\noindent
\textbf{Generating textual humor.} JAPE~\cite{JAPE} also uses phonological ambiguity to generate pun-based riddles. 
While our task involves producing free-form responses to a novel stimulus, JAPE produces stand-alone ``canned'' jokes. 
HAHAcronym~\cite{HAHAcronym} generates a funny expansion of a given acronym. Unlike our work, HAHAcronym operates on text, and is limited to producing sets of words. 
~\newcite{UnsupervisedJokeBigData} develop an unsupervised model that produces jokes of the form, ``\emph{I like my X like I like my Y, Z}'' .

\par \noindent
\textbf{Generating multi-modal humor.}~\newcite{meme} predict a meme's text based on a given funny image. Similarly,~\newcite{shahaf2015inside} and~\newcite{radev2015humor} learn to rank cartoon captions based on their funniness. Unlike typical, boring images in our task, memes and cartoons are images that are already funny or atypical. E.g., ``LOL-cats'' (funny cat photos), ``Bieber-memes'' (modified pictures of Justin Bieber), cartoons with talking animals, etc. 
~\newcite{chandrasekaran2015we} alter an abstract scene to make it more funny. In comparison, our task is to generate witty natural language remarks for a novel image. 

\par \noindent
\textbf{Poetry generation.} Although our tasks are different, our generation approach is conceptually similar to~\newcite{Ghazvininejad2016GeneratingTP} who produce poetry, given a topic.  While they also generate and score a set of candidates, their approach involves many more constraints and utilizes a finite state acceptor unlike our approach which enforces constraints during beam search of the RNN decoder.
\section{Approach}
\label{sec:approach}
 
\noindent
\textbf{Extracting tags}. 
The first step in producing a contextually witty remark is to identify concepts that are relevant to the context (image). 
At times, these concepts are directly available as e.g., tags posted on social media. 
We consider the general case where such tags are unavailable, and automatically extract tags associated with an image.

We extract the top-5 object categories predicted by a state-of-the-art Inception-ResNet-v2 model~\cite{szegedy2017inception} trained for image classification on ImageNet~\cite{deng2009imagenet}. 
We also consider the words from a (boring) image description (generated from~\newcite{vinyals2016show}). 
We combine the classifier object labels and words from the caption (ignoring stopwords) to produce a set of tags associated with an image, as shown in Fig.~\ref{fig:model}. 
We then identify concepts from this collection that can potentially induce wit.

\noindent
\textbf{Identifying puns}. 
We attempt to induce an incongruity by using a pun in the image description. 
We identify candidate words for linguistic wordplay by comparing image tags against a list of puns. 

We construct the list of puns by mining the web for differently spelled words that sound exactly the same (heterographic homophones). 
We increase coverage by also considering pairs of words with 0 edit-distance, according to a metric based on fine-grained articulatory representations (AR) of word pronunciations~\cite{jyothi2014revisiting}.
Our list of puns has a total of 1067 unique words (931 from the web and 136 from the AR-based model). 

The pun list yields a set of puns that are associated with a given image and their phonologically identical counterparts, which together form the \emph{pun vocabulary} for the image.  
We evaluate our approach on the subset of images that have non-empty pun vocabularies (about 2 in 5 images). 

\noindent
\textbf{Generating punny image captions}. 
We introduce an incongruity by forcing a vanilla image captioning model~\cite{vinyals2016show} to decode a \emph{phonological counterpart} of a pun word associated with the image, at a specific time-step during inference (e.g., `sell' or `sighed', showed in orange in Fig.~\ref{fig:model}). 
We achieve this by limiting the vocabulary of the decoder at that time-step to only contain counterparts of image-puns. 
In following time-steps, the decoder generates new words conditioned on all previously decoded words. 
Thus, the decoder attempts to generate sentences that flow well based on previously uttered words. 

We train two models that decode an image description in forward (start to end) and reverse (end to start) directions, depicted as `fRNN' and `rRNN' in Fig.~\ref{fig:model} respectively. The fRNN can decode words after accounting for the incongruity that occurs early in the sentence and the rRNN is able to decode the early words in the sentence after accounting for the incongruity that can occur later. 
The forward RNN and reverse RNN generate sentences in which the pun appears in each of the first $T$ and last $T$ positions, respectively.\footnote{For an image, we choose $T=\{1,2,...,5\}$ and beam size $= 6$ for each decoder. This generates a pool of $5$ (T) $*$ $6$ (beam size) $*$ $2$ (forward + reverse decoder) $=60$ candidates.
}

\noindent
\textbf{Retrieving punny image captions}. 
	As an alternative to our approach of \emph{generating} witty remarks for the given image, we also attempt to leverage natural, human-written sentences which are relevant (yet unexpected) in the given context. 
	Concretely, we retrieve natural language sentences\footnote{To prevent the context of the sentence from distracting the perceiver, we consider sentences with $< 15$ words. 
	Overall, we are left with a corpus of about 13.5 million sentences.} from a combination of the Book Corpus~\cite{zhu2015aligning} and corpora from the NLTK toolkit~\cite{Loper2002}. 
	The retrieved sentences each (a) contains an incongruity (pun) whose counterpart is associated with the image, and (b) has support in the image (contains an image tag). 
	This yields a pool of candidate captions that are perfectly grammatical, a little unexpected, and somewhat relevant to the image (see Sec.~\ref{subsec:qual}). 

\noindent
\textbf{Ranking}. 
	We rank captions in the candidate pools from both generation and retrieval models, according to their log-probability score under the image captioning model. 
	We observe that the higher-ranked descriptions are more relevant to the image and grammatically correct. 
	We then perform non-maximal suppression, i.e., eliminate captions that are similar\footnote{Two sentences are similar if the cosine similarity between the average of the Word2Vec~\cite{word2vec} representations of words in each sentence is $\geq$ 0.8. } to a higher-ranked caption to reduce the pool to a smaller, more diverse set. 
	We report results on the 3 top-ranked captions. 
	We describe the effect of design choices in the supplementary. 
\section{Results}

\noindent
\textbf{Data}. 
We evaluate witty captions from our approach via human studies. 100 random images (having associated puns) are sampled from the validation set of COCO~\cite{lin2014microsoft}. 

\noindent
\textbf{Baselines}. 
\label{subsec:baselines}
	We compare the wittiness of descriptions generated by our model against 3 qualitatively different baselines, and a human-written witty description of an image. Each of these evaluates a different component of our approach. 
	\textbf{Regular inference} generates a fluent caption that is relevant to the image but is not attempting to be witty. 
	\textbf{Witty mismatch} is a human-written witty caption, but for a different image from the one being evaluated. This baseline results in a caption that is intended to be witty, but does not attempt to be relevant to the image. 
	\textbf{Ambiguous} is a `punny' caption where a pun word in the boring (regular) caption is replaced by its counterpart. This caption is likely to contain content that is relevant to the image, \emph{and} it contains a pun. However, the pun is not being used in a fluent manner. 

	We evaluate the \textbf{image-relevance} of the top witty caption by comparing against a boring machine caption and a random caption (see supplementary).

\noindent
\textbf{Evaluating annotations}. 
	Our task is to generate captions that a layperson might find witty. 
	To evaluate performance on this task, we ask people on Amazon Mechanical Turk (AMT) to vote for the wittier among the given pair of captions for an image. We collect annotations from 9 unique workers for each relative choice and take the majority vote as ground-truth. 
	For each image, we compare each of the generated 3 top-ranked and 1 low-ranked caption against 3 baseline captions and 1 human-written witty caption.\footnote{This results in a total of $4$ (captions) $*2$ (generation $+$ retrieval) $*4$ (baselines $+$ human caption) $= 32$ comparisons of our approach against baselines. 
    We also compare the wittiness of the 4 generated captions against the 4 retrieved captions (see supplementary) for an image (16 comparisons). 
    In total, we perform 48 comparisons per image, for 100 images.}

\noindent
\textbf{Constrained human-written witty captions}. 
We evaluate the ability of humans and automatic methods to use the given context \emph{and pun words} to produce a caption that is perceived as witty. 
We ask subjects on AMT to describe a given image in a witty manner. 
To prevent observable \emph{structural} differences between machine and human-written captions, we ensure consistent pun vocabulary (utilization of pre-specified puns for a given image). We also ask people to avoid first person accounts or quote characters in the image. 

\begin{figure}[t!]
\centering 
\includegraphics[width=2.6in, height=1.7in]{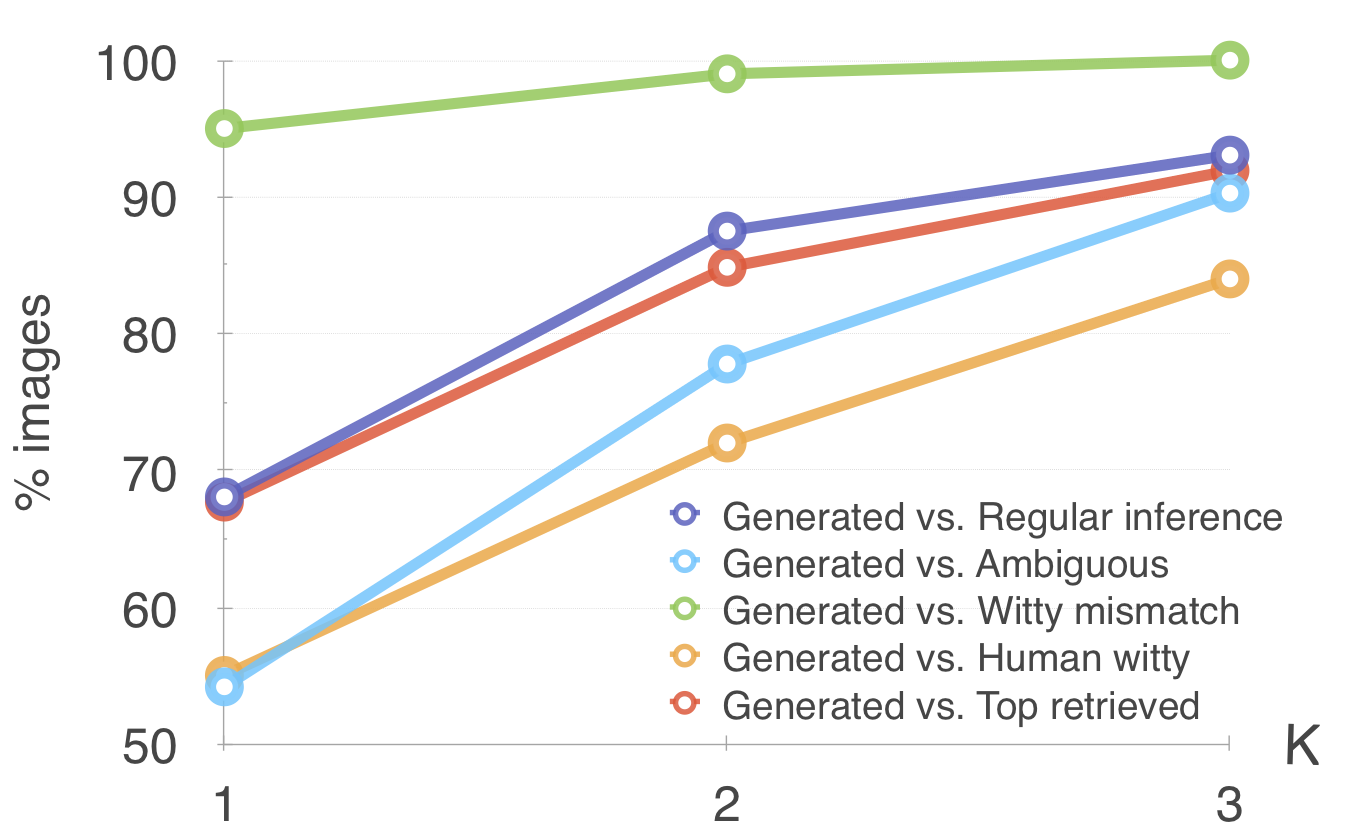}
    \caption{
    	\fontsize{10}{11}\selectfont{
        Wittiness of top-3 generated captions vs. other approaches. 
      y-axis measures the \% images for which at least one of $K$ captions from our approach is rated wittier than other approaches. 
      Recall steadily increases with the number of generated captions ($K$).
      }
      }
        \label{fig:recall_at_k}      
 \end{figure}

\noindent
\textbf{Metric}. 
~\ref{fig:recall_at_k}, we report performance of the generation approach using the Recall@K metric. For $K=1,2,3$, we plot the percentage of images for which at least one of the $K$ `best' descriptions from our model outperformed another approach. 
 
\begin{figure*}[t!]
\setlength{\fboxsep}{0pt}
\setlength{\fboxrule}{0pt}
	\begin{subfigure}[t]{1.44in}
 \fbox{\includegraphics[width=1.6in,height=1.2in]{figures/COCO_val2014_000000350988.jpg}}
		\caption{
        \textbf{Generated}: a bored (board) bench sits in front of a window. \\
        \textbf{Retrieved}: Wedge sits on the bench opposite Berry, bored (board). \\
        \textbf{Human}: could you please make your pleas (please)!
}		\label{fig:bored_bench}
	\end{subfigure}
    \hfill    
	\begin{subfigure}[t]{1.44in}
  \fbox{\includegraphics[width=1.6in,height=1.2in]{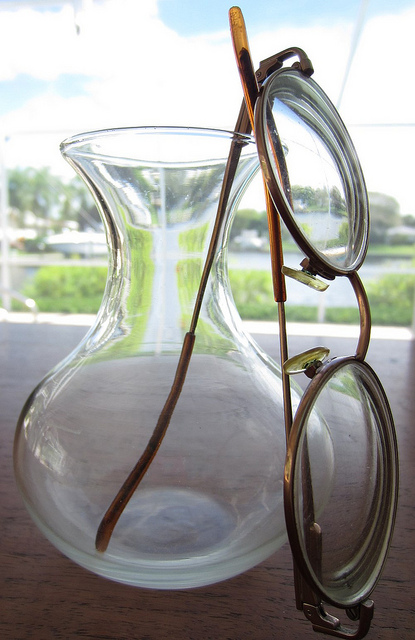}}
		\caption{
        \textbf{Generated}: a loop (loupe) of flowers in a glass vase. \\
        \textbf{Retrieved}: the flour (flower) inside teemed with worms. \\ 
        \textbf{Human}: piece required for peace (piece).
}		\label{fig:vase}
	\end{subfigure}
    \hfill
    \begin{subfigure}[t]{1.44in}
 \fbox{\includegraphics[width=1.6in,height=1.2in]{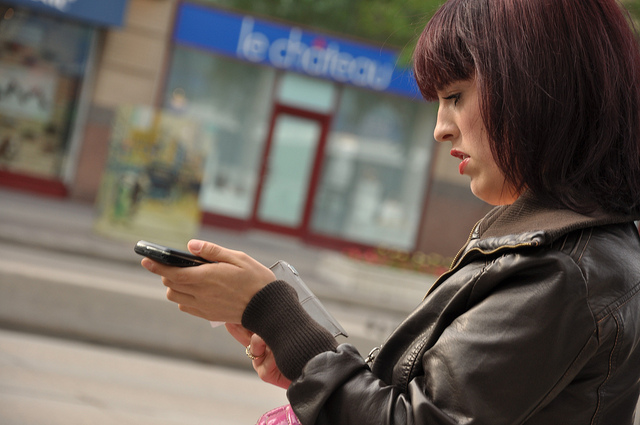}}
		\caption{\textbf{Generated}: a woman sell (cell) her cell phone in a city. \\
        \textbf{Retrieved}: Wright (right) slammed down the phone. \\
        \textbf{Human}: a woman sighed (side) as she regretted the sell.
}		\label{fig:woman}
	\end{subfigure}
    \hfill    
	\begin{subfigure}[t]{1.44in}
  \fbox{\includegraphics[width=1.45in,height=1.2in]{figures/COCO_val2014_000000430681.jpg}}
		\caption{
        \textbf{Generated}: a bear that is bare (bear) in the water. \\
        \textbf{Retrieved}: water glistened off her bare (bear) breast. \\
        \textbf{Human}: you won't hear a creak (creek) when the bear is feasting. 
}		\label{fig:water_bear}
	\end{subfigure}
    \\
    \begin{subfigure}[t]{1.44in}
 \fbox{\includegraphics[width=1.6in,height=1.2in]{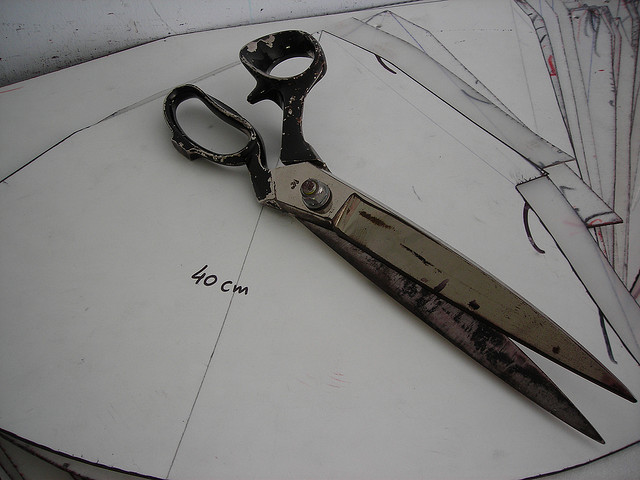}}
		\caption{
        \textbf{Generated}: a loop (loupe) of scissors and a pair of scissors. \\
        \textbf{Retrieved}: i continued slicing my pear (pair) on the cutting board. \\
        \textbf{Human}: the scissors were near, but not clothes (close).
}		\label{fig:scissors}
	\end{subfigure}
    \hfill    
	\begin{subfigure}[t]{1.44in}
  \fbox{\includegraphics[width=1.6in,height=1.2in]{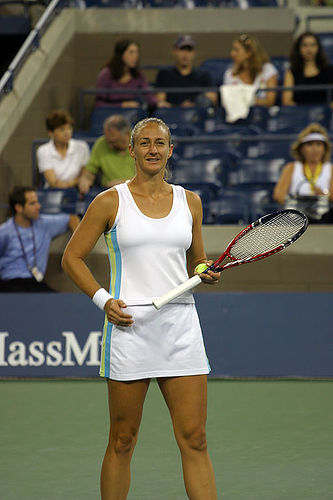}}
		\caption{
        \textbf{Generated}: a female tennis player caught (court) in mid swing. \\
        \textbf{Retrieved}: i caught (court) thieves on the roof top. \\ 
        \textbf{Human}: the man made a loud bawl (ball) when she threw the ball. 
}		\label{fig:tennis}
	\end{subfigure}
    \hfill
    \begin{subfigure}[t]{1.44in}
 \fbox{\includegraphics[width=1.6in,height=1.2in]{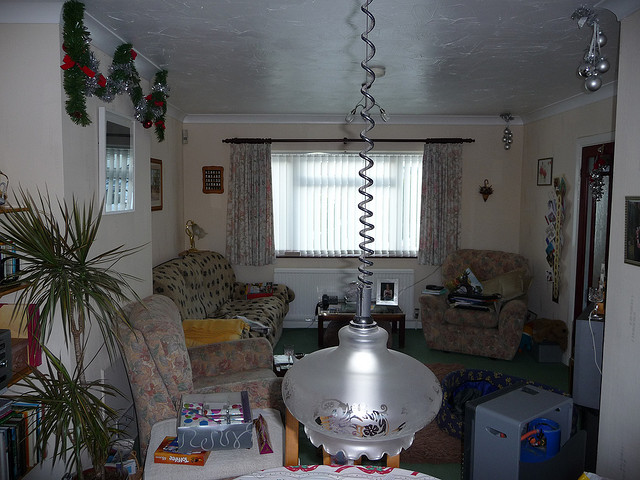}}
		\caption{
        	\textbf{Generated}: a bored (board) living room with a large window. \\
       	 \textbf{Retrieved}: anya sat on the couch, feeling bored (board). \\ 
         \textbf{Human}: the sealing (ceiling) on the envelope resembled that in the ceiling.
}		\label{fig:living_room}
	\end{subfigure}
    \hfill    
	\begin{subfigure}[t]{1.44in}
  \fbox{\includegraphics[width=1.45in,height=1.2in]{figures/COCO_val2014_000000568147.jpg}}
			\caption{
            	\textbf{Generated}: a parking meter with rode (road) in the background. \\ 
                \textbf{Retrieved}: smoke speaker sighed (side). \\ 
                \textbf{Human}: a nitting of color didn't make the poll (pole) less black.}
        \label{fig:parking_meter}
	\end{subfigure}
\caption{
	\fontsize{10}{11}\selectfont{
        	The top row contains selected examples of human-written witty captions, and witty captions generated and retrieved from our models. The examples in the bottom row are randomly picked. 
            }
	}
\label{fig:qual_egs}
\end{figure*}

\noindent
\textbf{Generated captions vs. baselines.} 
As we see in Fig.~\ref{fig:recall_at_k}, the top generated image description (top-1G) is perceived as wittier compared to all baseline approaches more often than not (the vote is $>$$50\%$ at $K=1$). 
We observe that as $K$ increases, the recall steadily increases, i.e., when we consider the top $K$ generated captions, increasingly often, humans find at least one of them to be wittier than captions produced by baseline approaches. 
People find the top-1G for a given image to be wittier than mismatched human-written image captions, about 95\% of the time. The top-1G is also wittier than a naive approach that introduces ambiguity about 54.2\% of the time. When compared to a typical, boring caption, the generated captions are wittier 68\% of the time. Further, in a head-to-head comparison, the generated captions are wittier than the retrieved captions 67.7\% of the time. 
We also validate our choice of ranking captions based on the image captioning model score. We observe that a `bad' caption, i.e., one ranked lower by our model, is significantly less witty than the top 3 output captions. 

Surprisingly, when the human is constrained to use the same words and style as the model, the generated descriptions from the model are found to be wittier for 55\% of the images. 
Note that in a Turing test, a machine would equal human performance at 50\%\footnote{Recall that this compares how a witty description is constructed, given the image \emph{and} specific pun words. A Turing test-style evaluation that compares the overall wittiness of a machine and a human would refrain from constraining the human in any way.}. 
This led us to speculate if the constraints placed on language and style might be restricting people's ability to be witty. We confirmed this by evaluating free-form human captions. 

\noindent
\textbf{Free-form Human-written Witty Captions.} 
\label{subsec:free_form}
	We ask people on AMT to describe an image (using any vocabulary) in a manner that would be perceived as funny. 
	As expected, when compared against automatic captions from our approach, human evaluators find free-form human captions to be wittier about 90\% of the time compared to 45\% in the case of constrained human witty captions. 
	Clearly, human-level creative language with unconstrained sentence length, style, choice of puns, etc., makes a significant difference in the wittiness of a description. 
	In contrast, our automatic approach is constrained by caption-like language, length, and a word-based pun list. Training models to intelligently navigate this creative freedom is an exciting open challenge.

\noindent
\textbf{Qualitative analysis}. 
\label{subsec:qual}
The generated witty captions exhibit interesting features like alliteration (`a bare black bear ...') in Fig.~\ref{fig:bear} and~\ref{fig:woman}. At times, both the original pun (pole) and its counterpart (poll) make sense for the image (Fig.~\ref{fig:knight}). Occasionally, a pun is naively replaced by its counterpart (Fig.~\ref{fig:bored_bench}) or rare puns are used (Fig.~\ref{fig:vase}). 
On the other hand, some descriptions (Fig.~\ref{fig:scissors} and~\ref{fig:parking_meter}) that are forced to utilize puns do not make sense. See supplementary for analysis of retrieval model.
\vspace{-10pt}
\section {Conclusion}
\vspace{-5pt}
	We presented novel computational models inspired by cognitive accounts to address the challenging task of producing contextually witty descriptions for a given image. 
	We evaluate the models via human-studies, in which they significantly outperform meaningful baseline approaches. 

\section*{Acknowledgements}
\vspace{-3pt}
\fontsize{10}{11}\selectfont{
	We thank Shubham Toshniwal for his advice regarding the automatic speech recognition model. 
    This work was supported in part by: 
    a NSF CAREER award, ONR YIP award, ONR Grant N00014-14-12713, PGA Family Foundation award, Google FRA, Amazon ARA, DARPA XAI grant to DP 
    and NVIDIA GPU donations, Google FRA, IBM Faculty Award, and Bloomberg Data Science Research Grant to MB.
}
\putbib
\end{bibunit}

\begin{thebibliography}{}
\expandafter\ifx\csname natexlab\endcsname\relax\def\natexlab#1{#1}\fi

\bibitem[{Attardo and Raskin(1991)}]{attardo1991script}
Salvatore Attardo and Victor Raskin. 1991.
\newblock Script theory revis (it) ed: Joke similarity and joke representation
  model.
\newblock {\em Humor-International Journal of Humor Research\/}
  4(3-4):293--348.

\bibitem[{Binsted(1996)}]{binsted1996machine}
Kim Binsted. 1996.
\newblock Machine humour: An implemented model of puns.
\newblock {\em PhD Thesis, University of Edinburgh\/} .

\bibitem[{Binsted and Ritchie(1997)}]{JAPE}
Kim Binsted and Graeme Ritchie. 1997.
\newblock Computational rules for generating punning riddles.
\newblock {\em Humor: International Journal of Humor Research\/} .

\bibitem[{Chandrasekaran et~al.(2016)Chandrasekaran, Kalyan, Antol, Bansal,
  Batra, Zitnick, and Parikh}]{chandrasekaran2015we}
Arjun Chandrasekaran, Ashwin Kalyan, Stanislaw Antol, Mohit Bansal, Dhruv
  Batra, C.~Lawrence Zitnick, and Devi Parikh. 2016.
\newblock We are humor beings: Understanding and predicting visual humor.
\newblock In {\em CVPR\/}.

\bibitem[{Deng et~al.(2009)Deng, Dong, Socher, Li, Li, and
  Fei-Fei}]{deng2009imagenet}
Jia Deng, Wei Dong, Richard Socher, Li-Jia Li, Kai Li, and Li~Fei-Fei. 2009.
\newblock Imagenet: A large-scale hierarchical image database.
\newblock In {\em Computer Vision and Pattern Recognition, 2009. CVPR 2009.
  IEEE Conference on\/}. IEEE, pages 248--255.

\bibitem[{Ghazvininejad et~al.(2016)Ghazvininejad, Shi, Choi, and
  Knight}]{Ghazvininejad2016GeneratingTP}
Marjan Ghazvininejad, Xing Shi, Yejin Choi, and Kevin Knight. 2016.
\newblock Generating topical poetry.
\newblock In {\em EMNLP\/}.

\bibitem[{Jaech et~al.(2016)Jaech, Koncel-Kedziorski, and
  Ostendorf}]{Jaech2016PhonologicalP}
Aaron Jaech, Rik Koncel-Kedziorski, and Mari Ostendorf. 2016.
\newblock Phonological pun-derstanding.
\newblock In {\em HLT-NAACL\/}.

\bibitem[{Jyothi and Livescu(2014)}]{jyothi2014revisiting}
Preethi Jyothi and Karen Livescu. 2014.
\newblock Revisiting word neighborhoods for speech recognition.
\newblock {\em ACL 2014\/} page~1.

\bibitem[{Kao et~al.(2013)Kao, Levy, and Goodman}]{Kao2013-kx}
Justine~T Kao, Roger Levy, and Noah~D Goodman. 2013.
\newblock The funny thing about incongruity: A computational model of humor in
  puns.
\newblock In {\em Proceedings of the Annual Meeting of the Cognitive Science
  Society\/}.

\bibitem[{Lin et~al.(2014)Lin, Maire, Belongie, Hays, Perona, Ramanan,
  Doll{\'a}r, and Zitnick}]{lin2014microsoft}
Tsung-Yi Lin, Michael Maire, Serge Belongie, James Hays, Pietro Perona, Deva
  Ramanan, Piotr Doll{\'a}r, and C~Lawrence Zitnick. 2014.
\newblock Microsoft coco: Common objects in context.
\newblock In {\em European Conference on Computer Vision\/}. Springer, pages
  740--755.

\bibitem[{Loper and Bird(2002)}]{Loper2002}
Edward Loper and Steven Bird. 2002.
\newblock Nltk: The natural language toolkit.
\newblock In {\em Proceedings of the ACL-02 Workshop on Effective Tools and
  Methodologies for Teaching Natural Language Processing and Computational
  Linguistics - Volume 1\/}. Association for Computational Linguistics, ETMTNLP
  '02.

\bibitem[{Mikolov et~al.(2013)Mikolov, Sutskever, Chen, Corrado, and
  Dean}]{word2vec}
Tomas Mikolov, Ilya Sutskever, Kai Chen, Greg~S Corrado, and Jeff Dean. 2013.
\newblock Distributed representations of words and phrases and their
  compositionality.
\newblock In {\em Advances in neural information processing systems\/}.

\bibitem[{Pepicello and Green(1984)}]{pepicello1984language}
William~J Pepicello and Thomas~A Green. 1984.
\newblock {\em Language of riddles: new perspectives\/}.
\newblock The Ohio State University Press.

\bibitem[{Petrovic and Matthews(2013)}]{UnsupervisedJokeBigData}
Sasa Petrovic and David Matthews. 2013.
\newblock Unsupervised joke generation from big data.
\newblock In {\em ACL\/}.

\bibitem[{Radev et~al.(2015)Radev, Stent, Tetreault, Pappu, Iliakopoulou,
  Chanfreau, de~Juan, Vallmitjana, Jaimes, Jha et~al.}]{radev2015humor}
Dragomir Radev, Amanda Stent, Joel Tetreault, Aasish Pappu, Aikaterini
  Iliakopoulou, Agustin Chanfreau, Paloma de~Juan, Jordi Vallmitjana, Alejandro
  Jaimes, Rahul Jha, et~al. 2015.
\newblock Humor in collective discourse: Unsupervised funniness detection in
  the new yorker cartoon caption contest.
\newblock {\em arXiv preprint arXiv:1506.08126\/} .

\bibitem[{Shahaf et~al.(2015)Shahaf, Horvitz, and Mankoff}]{shahaf2015inside}
Dafna Shahaf, Eric Horvitz, and Robert Mankoff. 2015.
\newblock Inside jokes: Identifying humorous cartoon captions.
\newblock In {\em Proceedings of the 21th ACM SIGKDD International Conference
  on Knowledge Discovery and Data Mining\/}. ACM, pages 1065--1074.

\bibitem[{Stock and Strapparava(2005)}]{HAHAcronym}
Oliviero Stock and Carlo Strapparava. 2005.
\newblock {HAHAcronym}: A computational humor system.
\newblock In {\em ACL\/}.

\bibitem[{Suls(1972)}]{TwoStageModelHumor}
Jerry~M Suls. 1972.
\newblock A two-stage model for the appreciation of jokes and cartoons: An
  information-processing analysis.
\newblock {\em The Psychology of Humor: Theoretical Perspectives and Empirical
  Issues\/} .

\bibitem[{Szegedy et~al.(2017)Szegedy, Ioffe, Vanhoucke, and
  Alemi}]{szegedy2017inception}
Christian Szegedy, Sergey Ioffe, Vincent Vanhoucke, and Alexander~A Alemi.
  2017.
\newblock Inception-v4, inception-resnet and the impact of residual connections
  on learning.
\newblock In {\em AAAI\/}. pages 4278--4284.

\bibitem[{Vinyals et~al.(2016)Vinyals, Toshev, Bengio, and
  Erhan}]{vinyals2016show}
Oriol Vinyals, Alexander Toshev, Samy Bengio, and Dumitru Erhan. 2016.
\newblock Show and tell: Lessons learned from the 2015 mscoco image captioning
  challenge.
\newblock {\em IEEE transactions on pattern analysis and machine
  intelligence\/} .

\bibitem[{Wang and Wen(2015)}]{meme}
William~Yang Wang and Miaomiao Wen. 2015.
\newblock I can has cheezburger? a nonparanormal approach to combining textual
  and visual information for predicting and generating popular meme
  descriptions.
\newblock In {\em NAACL\/}.

\bibitem[{Yu et~al.(2016)Yu, Nicolich-Henkin, Black, and
  Rudnicky}]{yu2016wizard}
Zhou Yu, Leah Nicolich-Henkin, Alan~W Black, and Alex~I Rudnicky. 2016.
\newblock A wizard-of-oz study on a non-task-oriented dialog systems that
  reacts to user engagement.
\newblock In {\em 17th Annual Meeting of the Special Interest Group on
  Discourse and Dialogue\/}. page~55.

\bibitem[{Zhu et~al.(2015)Zhu, Kiros, Zemel, Salakhutdinov, Urtasun, Torralba,
  and Fidler}]{zhu2015aligning}
Yukun Zhu, Ryan Kiros, Rich Zemel, Ruslan Salakhutdinov, Raquel Urtasun,
  Antonio Torralba, and Sanja Fidler. 2015.
\newblock Aligning books and movies: Towards story-like visual explanations by
  watching movies and reading books.
\newblock In {\em Proceedings of the IEEE International Conference on Computer
  Vision\/}. pages 19--27.

\bibitem[{Zwicky and Zwicky(1986)}]{zwicky1986imperfect}
Arnold Zwicky and Elizabeth Zwicky. 1986.
\newblock Imperfect puns, markedness, and phonological similarity: With fronds
  like these, who needs anemones.
\newblock {\em Folia Linguistica\/} 20(3-4):493--503.

\end{thebibliography}


\begin{thebibliography}{}
\expandafter\ifx\csname natexlab\endcsname\relax\def\natexlab#1{#1}\fi

\bibitem[{Jozefowicz et~al.(2016)Jozefowicz, Vinyals, Schuster, Shazeer, and
  Wu}]{jozefowicz2016exploring}
Rafal Jozefowicz, Oriol Vinyals, Mike Schuster, Noam Shazeer, and Yonghui Wu.
  2016.
\newblock Exploring the limits of language modeling.
\newblock {\em arXiv preprint arXiv:1602.02410\/} .

\bibitem[{Kiros et~al.(2014)Kiros, Salakhutdinov, and
  Zemel}]{kiros2014unifying}
Ryan Kiros, Ruslan Salakhutdinov, and Richard~S Zemel. 2014.
\newblock Unifying visual-semantic embeddings with multimodal neural language
  models.
\newblock {\em arXiv preprint arXiv:1411.2539\/} .

\bibitem[{Mikolov et~al.(2013)Mikolov, Sutskever, Chen, Corrado, and
  Dean}]{mikolov2013distributed}
Tomas Mikolov, Ilya Sutskever, Kai Chen, Greg~S Corrado, and Jeff Dean. 2013.
\newblock Distributed representations of words and phrases and their
  compositionality.
\newblock In {\em Advances in neural information processing systems\/}.

\end{thebibliography}

\appendix
\begin{bibunit}
\vspace{15pt}

\noindent
{\fontsize{12}{12}\selectfont \textbf{Appendix}} \par
\vspace{5pt}
\noindent
We first present details and results for the additional experiments that evaluate the relevance of the top generated caption from the model to the image, and compare the retrieved captions to baseline approaches. We then briefly discuss the rationales for the design choices that we made in our approach. Subsequently, we briefly describe some characteristics of pun words and follow it up with brief qualitative analysis of the output from the retrieval model. Finally, we present the annotation and evaluation interfaces that the human subjects interacted with. 

\section{Additional Experiments}

\subsection{Relevance of witty caption to image} 
We compared the relative relevance of the top witty caption from our generation approach against a machine generated boring caption (either for the same image or for a different, randomly chosen image) in a pairwise comparison. We showed Turkers an image and a pair of captions, and asked them to choose the more relevant caption for the image. We see that on average, the generated witty caption is considered more relevant than a machine generated boring caption for the same image 37.5\% of the time. People found the generated witty caption to be more relevant than a random caption 97.2\% of the time. This shows that in an effort to generate witty content, our approach produces descriptions that are a little less relevant compared to a boring description for the image. But our witty caption is clearly still relevant to the image (almost always more relevant than an unrelated caption).

\begin{figure}[t!]
	\includegraphics[width=3.45in, height=2.25in]{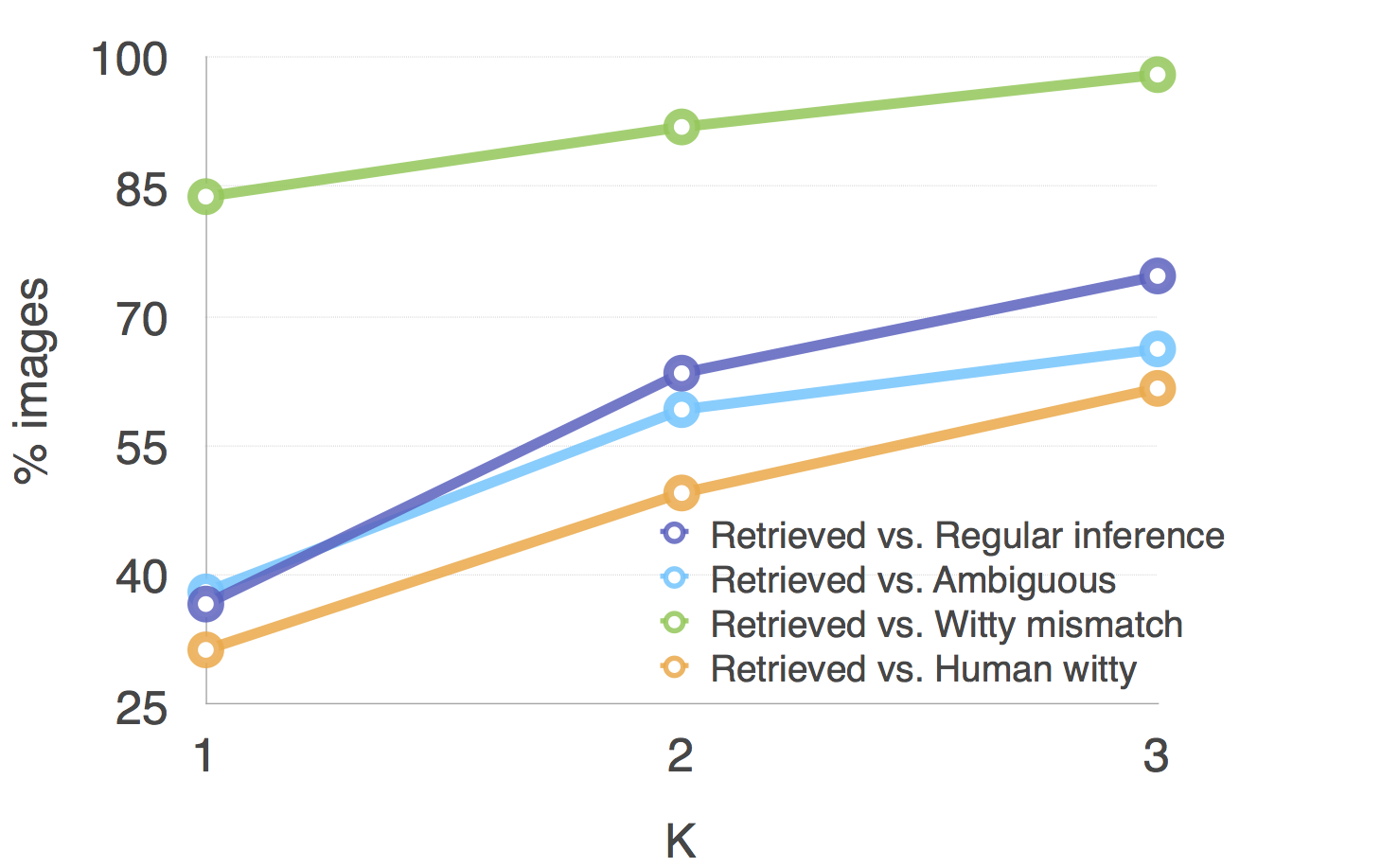}
    \caption{
        Comparison of wittiness of the top 3 captions from our retrieval approach vs. other approaches. 
      The y-axis measures the \% images for which at least one of $K$ captions from our approach is rated wittier than other approaches. 
      As we increase the number of retrieved captions ($K$), recall steadily increases.}
    \label{fig:ret_recall_at_k}
 \end{figure}

\noindent
\subsection{Retrieved captions vs. baselines} 
Humans evaluate the wittiness of each of the 3 top-ranked retrieved captions against baseline approaches and a human witty caption. As we see in Fig.~\ref{fig:ret_recall_at_k}, at $K=1$, the top  retrieved description is found to be wittier than only a human-written witty caption that is mismatched with the given image (witty mismatch) 83.8\% of the time. The top retrieved caption is found \emph{less} witty than even a typical caption (regular inference) about 63.4\% of the time. Similarly, the retrieved caption is also found to be less witty than a naive method that produces punny captions (ambiguous) about 62\% of the time. We observe the trend that as $K$ increases, recall also increases. On average, at least one of the top 3 retrieved captions is wittier than the (constrained) human witty caption about 61.6\% of the time, compared to generated captions which are wittier 84.0\% of the time.  

Poor performance of retrieved captions could be due to the fact that they are often not perfectly apt for the given image since they are retrieved from story-based corpora. Please see Sec.~\ref{subsec:qual} for examples, and a more detailed discussion. 
As we will see in the next section, these issues do not extend to the generation approach which exhibits strong performance against baseline approaches, human-written witty captions and the retrieval approach. 
While these captions might evoke a sense of incongruity, it is likely hard for the viewer to resolve the alternate interpretation of the retrieved caption as being applicable to the image.
 
\section{Design choices}
In this section, we describe how our architecture design and parameter choices in the architectures influence witty descriptions. 
During the design of our model, we made choices of parameters based on observations from qualitative results. 
For instance, we experimented with different beam sizes to generate a set of high precision captions with few false positives. We found that a beam size of 6 resulted in a sufficient number of candidate sentences which were reasonably accurate. 
We extract image tags from the top-K predictions of an image classifier. We experimented with different values of K, where $K \in \{1,5,10\}$. We also tried using a score threshold, where classes predicted with a score above the threshold were considered valid image tags. We found that $K=5$ results in reasonable predictions. Determining a reasonable threshold on the other hand was difficult because for most images, class prediction scores are extremely peaky. 
We also experimented with the different positions that a pun counterpart can be forced to appear in. Based on qualitative examples, we found that the model generated witty descriptions that were somewhat sensible when a pun word appeared at any of the first or last 5 positions of a sentence. We also experimented with a number of different methods to re-rank the candidate of witty captions, e.g., language model score~\cite{jozefowicz2016exploring}, image-sentence similarity score~\cite{kiros2014unifying}, semantic similarity (using Word2Vec~\cite{mikolov2013distributed}) of the pun counterpart to the sentence, a priori probability of the pun counterpart in a large corpus of English sentences to avoid rare / unfamiliar words, likelihood of the tag (under the image captioning model or the classifier as applicable). etc. We qualitatively found that re-ranking using log. prob. score of the image captioning model, while being the simplest, resulted in the best set of candidate witty captions. 

\section{Pun List}
Recall that we construct a list of puns by mining the web and based on automatic methods that measure the similarity of pronunciation of words. 
Upon inspecting our list of puns, we observe that it contains puns of many frequently used words and some pun words that are rarely used in everyday language, e.g., `wight' (which is the counterpart of `white'). 
Since a rare pun word can be distracting to a perceiver, the corresponding caption might be harder to resolve, making it less likely to be perceived as witty. 
Thus, we see limited benefit in increasing the size of our pun list further to include words that are used even less frequently.

\section{Qualitative analysis of retrieved descriptions}
\label{subsec:qual}

The retrieved witty descriptions are retrieved from story-based corpora. They often contain sentences that describe a very specific situation or instance. Although these sentences are grounded in objects that are also present in the image, the entire sentence often contains a few words that are irrelevant for a given image, as we see in Fig.~\ref{fig:bored_bench}, Fig.~\ref{fig:vase} and Fig.~\ref{fig:tennis}. This is a likely reason for why a retrieved sentence containing a pun is perceived as less witty when compared with witty descriptions generated for the image. 
 
\begin{figure*}[t!]
\centering
\setlength{\fboxsep}{0pt}
\setlength{\fboxrule}{0pt}
	\begin{subfigure}[t]{2.1in}
  \fbox{\includegraphics[width=2.1in, height=1.8in]{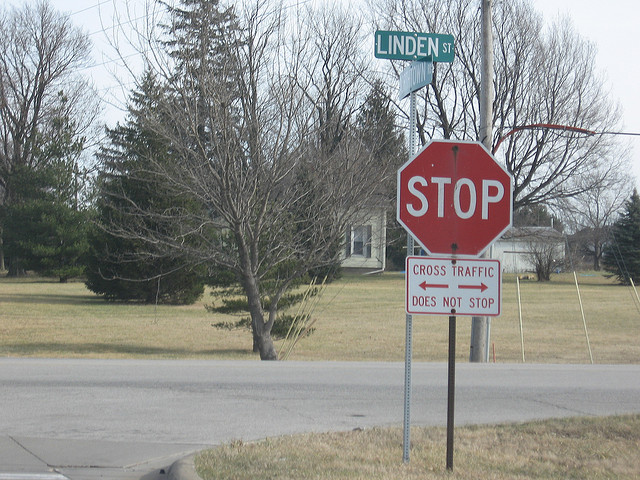}}
		\caption{
        \textbf{Generated}: a read (red) stop sign on a street corner. \\
        \textbf{Retrieved}: the sign read (red) : \\
        \textbf{Human}: i sighed (side) when we got off the main street. 
}		\label{fig:road_sign}
		\end{subfigure}	
	\hspace{0.05in}    
    \begin{subfigure}[t]{2.1in}
  \fbox{\includegraphics[width=2.1in, height=1.8in]{figures/COCO_val2014_000000350988.jpg}}
		\caption{
        \textbf{Generated}: a bored (board) bench sits in front of a window. \\
        \textbf{Retrieved}: Wedge sits on the bench opposite Berry, bored (board). \\
        \textbf{Human}: could you please make your pleas (please)!
}		\label{fig:bored_bench}
		\end{subfigure}  	
	\hspace{0.05in}    
	\begin{subfigure}[t]{2.1in}
  		\fbox{\includegraphics[width=2.1in, height=2.25in]{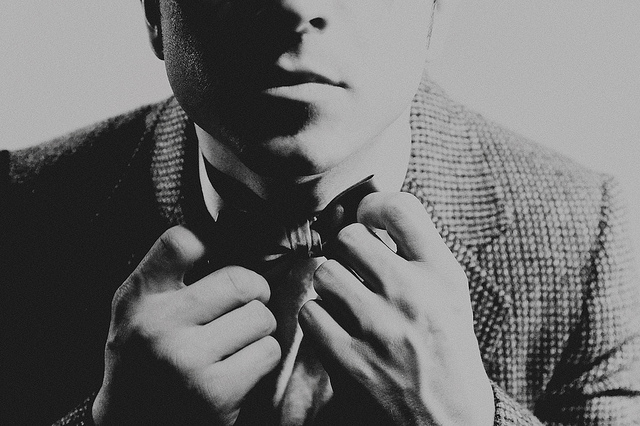}}
		\caption{\textbf{Generated}: a man wearing a loop (loupe) and a bow tie. \\
        \textbf{Retrieved}: meanwhile the man's head turned back to beau (bow) with his eyes wide. \\
        \textbf{Human}: he could be my beau (bow) with that style.
}		\label{fig:woman}
	\end{subfigure}
    \hspace{0.05in}    
	\begin{subfigure}[t]{2.1in}
  \fbox{\includegraphics[width=2.1in, height=2.25in]{figures/COCO_val2014_000000471691.jpg}}
		\caption{
        \textbf{Generated}: a loop (loupe) of flowers in a glass vase. \\
        \textbf{Retrieved}: the flour (flower) inside teemed with worms. \\ 
        \textbf{Human}: piece required for peace (piece).
}		\label{fig:vase}
	\end{subfigure}
	\hspace{0.05in}    
	\begin{subfigure}[t]{2.1in}
  		\fbox{\includegraphics[width=2.1in, height=2.25in]{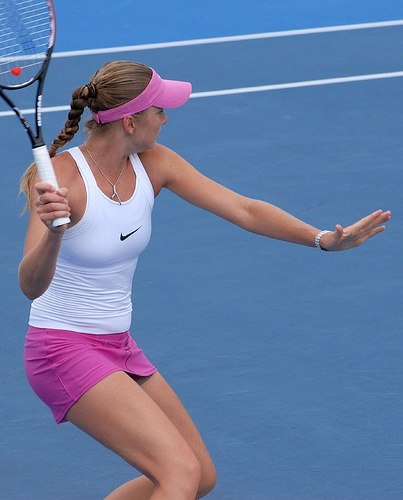}}
		\caption{
        \textbf{Generated}: a female tennis player caught (court) in mid swing. \\
        \textbf{Retrieved}: my shirt caught (court) fire. \\
        \textbf{Human}: the woman's hand caught (court) in the center.
}		\label{fig:tennis}
	\end{subfigure}
	\hspace{0.05in}    
	\begin{subfigure}[t]{2.1in}
  		\fbox{\includegraphics[width=2.1in, height=2.25in]{figures/COCO_val2014_000000559707.jpg}}
		\caption{
        \textbf{Generated}: broccoli and meet (meat) on a plate with a fork. \\
        \textbf{Retrieved}: ``you mean white folk (fork)''. \\
        \textbf{Human}: the folk (fork) enjoyed the food with a fork.
}		\label{fig:broccoli}
	\end{subfigure}            
\caption{
Sample images and witty descriptions from our generation model, retrieval model and a human. The puns (counterparts) that are used in captions (a) to (f) are read/sighed, bored, loop/beau, loop/flour/peace, caught and meet/folk respectively. The word in the parenthesis  following each counterpart is the pun associated with the image. It is provided as a reference to the source of the unexpected pun which is used in the caption.
}
\label{fig:qual_egs}
\end{figure*} 

\section{Interface for `Be Witty!'}
\begin{figure*}
    \centering
    \includegraphics[scale=0.4]{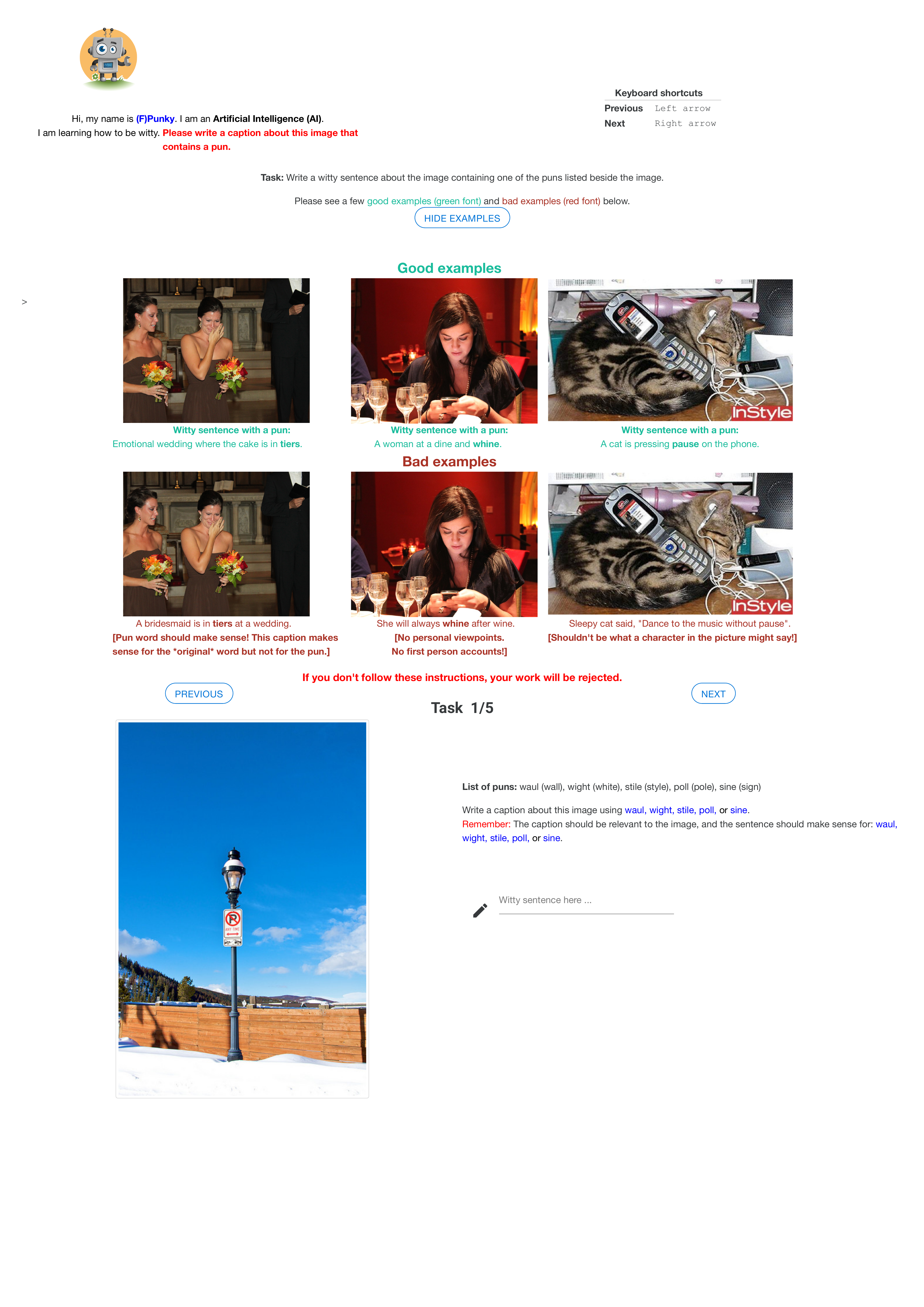}
    \vspace{-100pt}
    \caption{AMT web interface for the `Be Witty!' task.}
    \label{fig:be_witty}
\end{figure*}

We ask people on Amazon Mechanical Turk (AMT) to create witty descriptions for the given image. We also ask them to utilize one of the given pun words associated with the image. We show them a few good and bad examples to illustrate the task better. Fig.~\ref{fig:be_witty} shows the interface that we used to collect these human-written witty descriptions for an image.

\section{Interface for `Which is wittier?'}
\begin{figure*}[t]
    \centering
    \includegraphics[scale=0.8]{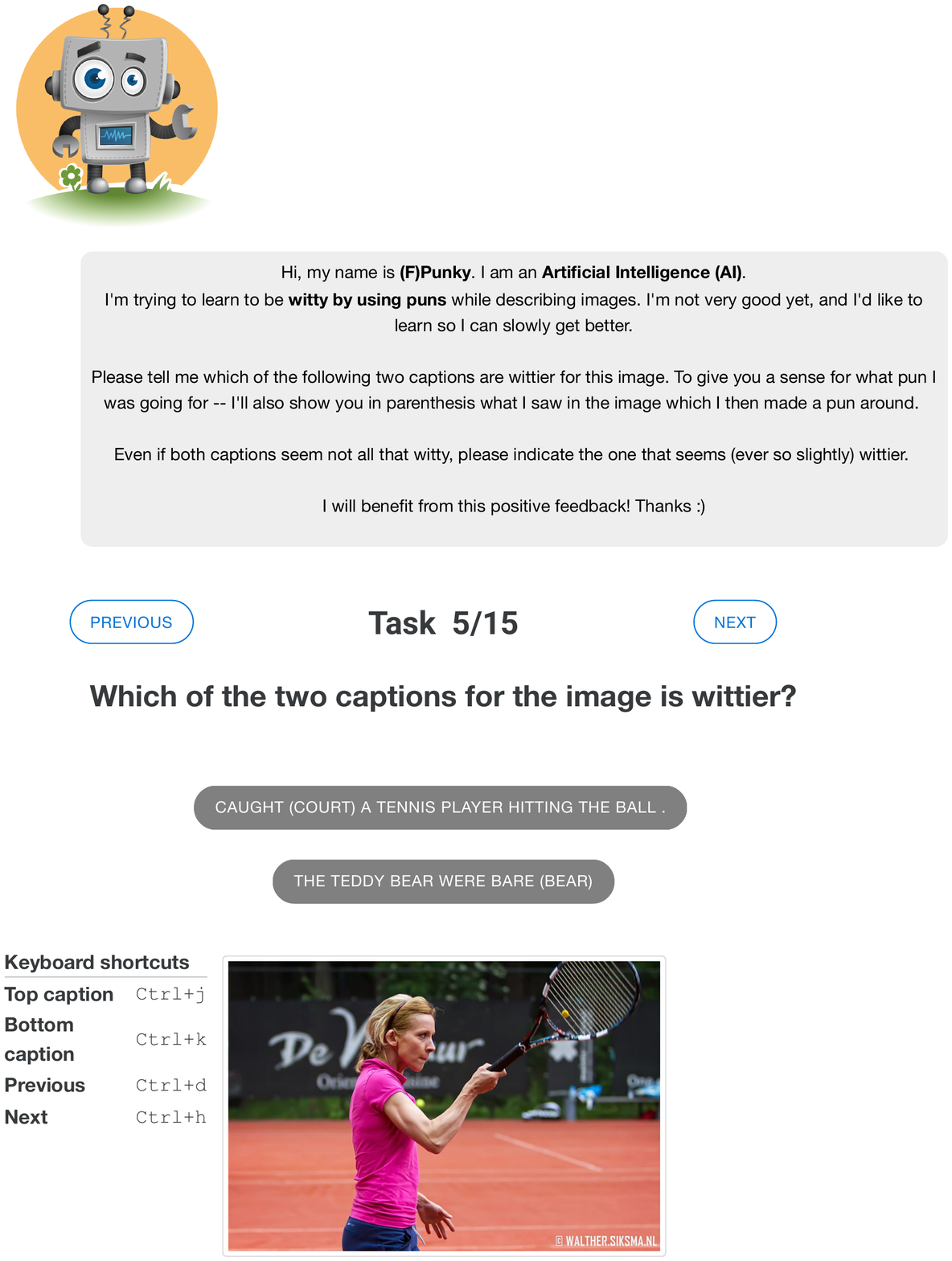}
    \vspace{-75pt}    
    \caption{AMT web interface for the `Which is wittier?' task.}
    \label{fig:which_is_witty}
\end{figure*}

We showed people on AMT two descriptions for a given image and asked them to click on the description that was wittier for the image. The web interface that we used to collect this data is shown in Fig.~\ref{fig:which_is_witty}.

\putbib
\end{bibunit}

\end{document}